\documentclass[nohyperref]{article}

\usepackage{microtype}
\usepackage{graphicx}
\usepackage{subfigure}
\usepackage{booktabs} %

\usepackage{hyperref}

\usepackage[accepted]{icml2022}

\usepackage{amsmath}
\usepackage{amssymb}
\usepackage{mathtools}
\usepackage{amsthm}

\usepackage[capitalize,noabbrev]{cleveref}

\usepackage{multirow}
\usepackage{array}
\usepackage{footnote}
\usepackage{color}
\usepackage{wrapfig}
\usepackage{makecell}

\theoremstyle{plain}

\theoremstyle{definition}

\theoremstyle{remark}

\newcommand{\tabincell}[2]{\begin{tabular}{@{}#1@{}}#2\end{tabular}}

\usepackage[textsize=tiny]{todonotes}
\usepackage[normalem]{ulem}

\begin{document}

\twocolumn[
\icmltitle{
HyperPELT: Unified Parameter-Efficient Language Model Tuning for Both Language and Vision-and-Language Tasks
}

\icmlsetsymbol{equal}{*}

\begin{icmlauthorlist}
\icmlauthor{Zhengkun Zhang}{equal,intern,nankai}
\icmlauthor{Wenya Guo}{equal,nankai}
\icmlauthor{Xiaojun Meng}{equal,huawei}
\icmlauthor{Yasheng Wang}{huawei}
\icmlauthor{Yadao Wang}{huawei}

\icmlauthor{Xin Jiang}{huawei}
\icmlauthor{Qun Liu}{huawei}
\icmlauthor{Zhenglu Yang}{nankai}
\end{icmlauthorlist}

\icmlaffiliation{nankai}{TKLNDST, CS, Nankai University, China}
\icmlaffiliation{huawei}{Noah's Ark Lab, Huawei Technologies}
\icmlaffiliation{intern}{Work is done at the internship of Noah's Ark Lab, Huawei Technologies.}

\icmlcorrespondingauthor{Zhenglu Yang}{yangzl@nankai.edu.cn}

\icmlkeywords{Machine Learning, ICML}

\vskip 0.4in
]

\printAffiliationsAndNotice{\icmlEqualContribution} %

\begin{abstract}
The workflow of pretraining and fine-tuning has emerged as a popular paradigm for solving various NLP and V\&L (Vision-and-Language) downstream tasks. With the capacity of pretrained models growing rapidly, how to perform parameter-efficient fine-tuning has become fairly important for quick transfer learning and deployment. In this paper, we design a novel unified parameter-efficient transfer learning framework that works effectively on both pure language and V\&L tasks. In particular, we use a shared hypernetwork that takes trainable hyper-embeddings as input, and outputs weights for fine-tuning different small modules in a pretrained language model, such as tuning the parameters inserted into multi-head attention blocks (\emph{i.e.,} prefix-tuning) and
feed-forward blocks (\emph{i.e.,} adapter-tuning). We define a set of embeddings (\emph{e.g.,} layer, block, task and visual embeddings) as the key components to calculate hyper-embeddings, which thus can support both  pure language and V\&L tasks. Our proposed framework adds fewer trainable parameters in multi-task learning while achieves superior performances and transfer ability compared to state-of-the-art methods. Empirical results on the GLUE benchmark and multiple V\&L tasks confirm the effectiveness of our framework on both textual and visual modalities.
\footnote{We will release our code to facilitate future work.}

\end{abstract}

\section{Introduction}
Pretraining and fine-tuning are now the prevalent paradigm in natural language processing, yielding state-of-the-art performances on a variety of downsteam tasks \cite{DBLP:conf/naacl/DevlinCLT19}. With pre-trained language models (PLMs) growing rapidly in size, it becomes increasingly infeasible to perform conventional fine-tuning on all model parameters, \emph{i.e., full fine-tuning}. 
It is even more time \& space-consuming for multi-tasking
if separate replicas of model parameters are updated and saved per single task.

To mitigate these issues, there has recently been one line of research on \textbf{P}arameter-\textbf{E}fficient \textbf{L}anguage model \textbf{T}uning (\textbf{PELT}).
A few lightweight transfer learning methods have been proposed and they only update a subset of model parameters while freeze the remaining most parameters \cite{liu2021autofreeze}.
Extra trainable task-specific model parameters can also be newly introduced to PLMs, such as the widely used adapter-tuning \cite{DBLP:conf/icml/HoulsbyGJMLGAG19} and prefix-tuning \cite{DBLP:conf/acl/LiL20} methods.
The former adapter-tuning adds new parameters between transformer layers, while the later prepends tunable prefix vectors to the keys and values of multi-head attention at each layer.
Although the number of parameters in the introduced adapter or prefix is much fewer than the original PLM, training these new parameters still requires a lot of resources due to the complex structure of PLMs.

Apart from traditional NLP tasks, fine-tuning language models pretrained on pure text corpora to perform various V\&L tasks, has merged as a upward trend.
Previous methods (\emph{e.g., VL-T5} from \citet{DBLP:conf/icml/ChoLTB21}) often concatenate visual patch tokens and textual tokens as input to a pretrained language model (\emph{e.g., T5} from \citet{DBLP:journals/jmlr/RaffelSRLNMZLL20}), and then finetune the whole model on V\&L tasks.
This tuning towards vision-and-language has achieved a noticeable improvement to V\&L tasks \cite{DBLP:conf/icml/ChoLTB21}.
The key advantage therein is that language models with large capacity and semantic interpretation serve as a cornerstone to benefit visual language alignment and modelling in a wide range of V\&L tasks. 

Similarly, training all the parameters of PLMs for handling visual input is time-consuming. It is crucial to explore how a small number of trainable parameters can equip a language model with the ability of handling visual input and V\&L tasks. Existing methods typically handle the visual input via a prompt-tuning form, and prepend visual patch tokens (\emph{i.e.,} visual prefix of \textit{Frozen} in~\citet{DBLP:journals/corr/abs-2106-13884}) to the textual sequence.
To reduce the trainable parameters, \emph{VL-adapter} \cite{sung2021vladapter} adopts the adapter-tuning technique from NLP to the frozen model \emph{VL-T5}, which can match the performance of \textit{full fine-tuning}.

Inspired by the recent progress of parameter-efficient tuning, we are motivated to unify a transfer learning framework that supports both language and V\&L models in tackling with multitasks.
We use a shared hypernetwork \cite{DBLP:conf/acl/MahabadiR0H20} that is able to take 
multi-task and multi-modal information
as input, and generate weights for tuning different task-specific modules of PLMs in transfer learning.
As shown in Figure~\ref{fig:method}, when finetuning on multitasks,
only the shared hypernetwork and its input embedding (namely, hyper-embedding) consisting of layer, block, task and visual embeddings, along with layer normalization, are trained.
Such unified parameter-efficient tuning reduces a great number of trainable parameters.

We experiment with two task-specific modules that use the weights output by our hypernetwork.
They are respectively multi-head attention modules \cite{DBLP:conf/acl/LiL20} and task-specific adapter \cite{DBLP:conf/icml/HoulsbyGJMLGAG19}.
Different from previous methods using visual input in a prompt-tuning manner, we present a novel perspective of adopting visual input to the above prefix-tuning and adapter-tuning modules.
Empirical results on GLUE benchmark and multiple V\&L tasks confirm the effectiveness of our unified framework.

In summary, we make the following contributions:
\begin{itemize} %
\setlength{\itemsep}{1pt}
\setlength{\parsep}{1pt}
\setlength{\parskip}{1pt}
\item We propose an unified parameter-efficient framework for vision and language transfer learning, which supports tuning both language and V\&L models on multitasks.
\item We present a novel method of leveraging visual modality as input for a shared hypernetwork, which generates weights for prefix-tuning and adapter-tuning modules.
\item We demonstrate that our framework scales more efficiently than prior work.
Empirical results on GLUE benchmark show the effectiveness of our proposed framework in multi-task learning. Empirical results on multiple vision-and-language tasks evidence its feasibility and usefulness in multi-modal transfer learning.
\item We also perform extensive experiments on few-shot domain transfer
in pure language and V\&L scenarios,
and results reveal that the learned shared knowledge across multitasks in our framework is able to positively transfer to unseen domain tasks.
\end{itemize}

\section{Related Work}

In this section, we review recent research on parameter-efficient tuning for pure language and V\&L tasks, as well as the corresponding work for multi-task learning.

\subsection{Parameter-efficient tuning}
As recent models grow rapidly in size, how to finetune pretrained models with a small number of trainable parameters becomes more crucial. Existing research \cite{DBLP:journals/corr/abs-2110-04366, DBLP:journals/corr/abs-2104-08691, DBLP:journals/corr/abs-2107-13586, DBLP:journals/corr/abs-2110-07577} have explored a large amount of methods on parameter-efficient tuning. These methods generally include two categories according to whether new trainable parameters are introduced. One category is that only a subset of model parameters can be updated while freezing the remain \cite{liu2021autofreeze, lee2019would}. The other is introducing a few task-specific new parameters to different parts of pretrained models, such as before multi-head attention \cite{DBLP:conf/acl/LiL20}, after feedforward layers \cite{DBLP:conf/icml/HoulsbyGJMLGAG19} or Mixed-and-Match methods (\textit{MAM adapter}) proposed by \citet{DBLP:journals/corr/abs-2110-04366}.

\subsection{Tuning towards Vision-and-Language}
In addition, fine-tuning language models pretrained on pure large text corpora have led to noticeable improvements to V\&L tasks. This line of research such as \emph{VL-T5} \cite{DBLP:conf/icml/ChoLTB21} and \emph{Frozen} \cite{DBLP:journals/corr/abs-2106-13884} attempts to tune large language models (e.g. \emph{T5}; \emph{GPT-3}) to achieve transfer learning for V\&L tasks. For example, \emph{Frozen} aligns the image representation into the word representation space of frozen \emph{GPT-3} model which thus is able to generate captions for those images. \emph{PICa} \cite{DBLP:journals/corr/abs-2109-05014} utilizes a pretrained image captioner to convert the image into captions that \emph{GPT-3} can understand, and then adapt \emph{GPT-3} to solve the VQA tasks in a few-shot manner. \citet{sung2021vladapter} introduces a limited set of new trainable parameters to \emph{VL-T5} via a adapter-based method that can match the performance of fine-tuning the entire model.

\subsection{Multi-task Learning}
Learning a unified model to perform well on multiple different tasks (\emph{i.e.,}  multi-task learning) is a challenging problem in both NLP and V\&L domains. It has to address many challenges such as catastrophic forgetting, and model overfitting in low-resource tasks while underfitting in high-resource tasks \cite{aharoni-etal-2019-massively}. \citet{radford2019language} highlights the ability of language models to perform a wide range of multitasks in a zero-shot setting. As mentioned above, involving task-specific new parameters such as adapter \cite{DBLP:conf/icml/HoulsbyGJMLGAG19}, can be trained for each task separately while keeping the model fixed. \citet{DBLP:conf/iclr/OswaldHSG20} propose a task-conditioned hypernetwork to generate all the weights for the targeted model, while \citet{DBLP:conf/acl/MahabadiR0H20} use a shared hypernetwork to only generate weights for a small number of parameters in adapter modules, to allow the model to adapt to each individual task efficiently.

\paragraph{Our motivation.} Different from mainstream V\&L models that append image tokens to the input sequence, we present a novel perspective of merging textual and visual modalities, by using image embedding and task-specific type embedding of multitasks as input to a shared hypernetwork, which generates weights for prefix-tuning and adpater-tuning modules of PLMs. At the same time, we notice a recent paper~\cite{DBLP:journals/corr/abs-2203-00759} that was publicly available days ago. This concurrent work shares the similar motivation like us on generating weights for prefix-tuning modules via a hypernetwork, but their method is only targeted at pure language tasks. Our unified framework is able to improve transfer learning in both pure text and vision-to-language multitasks, in a very parameter-efficient manner.

\begin{figure*}
\centering
\includegraphics[width=\textwidth]{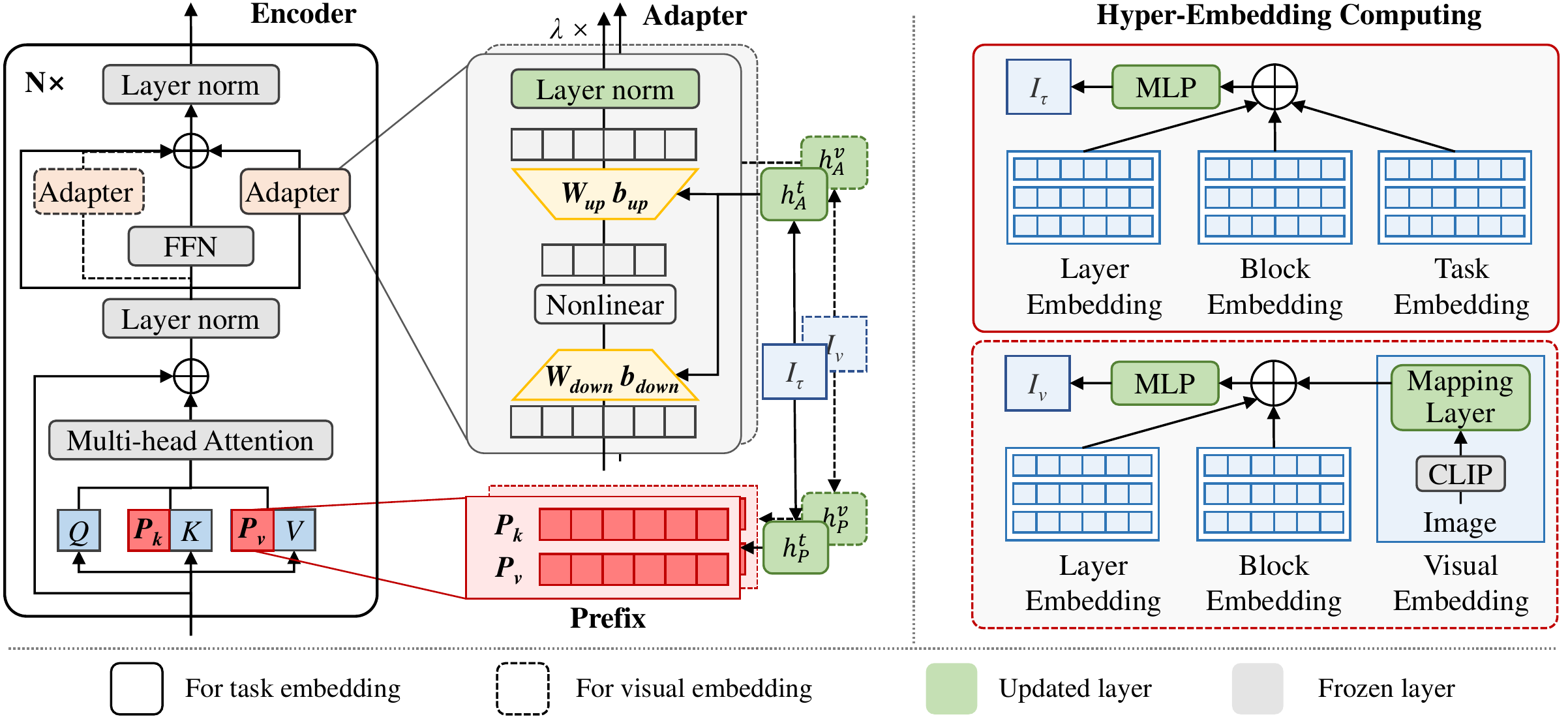}
\caption{
The model structure of the proposed unified pure language and V\&L multi-task framework (left), and illustration of computing the hyper-embedding (right). We use green color to fill the trainable layers and blue color for the frozen ones.
And the dashed parts denote the modules for processing visual modality.
Conditioned on the input hyper-embedding $I_{\tau}$ or $I_v$, 
the adapter hypernetwork $h^t_A$ or $h^v_A$ parallelly produce the weights ($W_{up}$, $b_{up}$, $W_{down}$, $b_{down}$) for  task-specific and visual-specific adapter-tuning modules. 
Similarly, the prefix hypernetwork $h^t_P$ or $h^v_P$ produce the weights ($P_k$, $P_v$) as the prefix-tuning vectors in multi-head attention modules.
During training, we only update layer normalization in \textit{T5}, the hypernetworks, and the used input embeddings (\textit{i.e.,} Layer, Block, Task and Visual).
}
\label{fig:method}
\end{figure*}

\section{Preliminaries}

\subsection{Pretrained Language Models}

All of our models are built on top of the state-of-the-art language model, \textit{T5}~\cite{DBLP:journals/jmlr/RaffelSRLNMZLL20},
consisting of an encoder-decoder Transformer~\cite{DBLP:conf/nips/VaswaniSPUJGKP17} with minor modifications. It frames
language tasks as sequence-to-sequence generation, and is trained simultaneously on multiple task datasets.
This large-scale \textit{T5} model achieves state-of-the-art performances across a diverse set of tasks. We use the \textit{T5} backbone as it enables training a universal model that interfaces with many downstream language tasks.

\subsection{Mutli-task Learning Problem formulation}
Our paper targets at a general multi-task learning problem, where we are given the data from a set of tasks $\{\mathcal{D}_{\tau}\}^T_{\tau=1}$. $T$ is the total number of tasks and $\mathcal{D}_{\tau}=\{(x^i_{\tau},y^i_{\tau})\}^{N_{\tau}}_{i=1}$ is the training data of the $\tau$-th task with $N_{\tau}$ samples.
We are also given a large-scale pretrained language model,
i.e., \textit{T5}, parameterized by $\theta$, which generates the output $y^i_{\tau}$ for input $x^i_{\tau}$.
The standard multi-task finetuning minimizes the following loss on the training set:
\begin{equation}
\mathcal{L}_{\text{total}} = \sum^T_{\tau=1} \sum_{(x^i_{\tau},y^i_{\tau})\in\mathcal{D}_{\tau}}
\mathcal{L}_{\text{task}} (\theta, x^i_{\tau}, y^i_{\tau}) ,
\label{equ:loss_1}
\end{equation}
where $\mathcal{L}_{\text{task}}$ is the loss function of the tasks that is usually defined as the cross-entropy loss.
Our goal is to efficiently finetune the given model in this multi-task learning setting, allowing knowledge sharing across tasks, and at the same time, enabling the model to adapt to each individual task.

\subsection{Hypernetworks}

The key idea of
hypernetwork~\cite{DBLP:conf/iclr/HaDL17,DBLP:conf/iclr/OswaldHSG20}
is to learn a parametric task-specific hyper-embedding $\{I_{\tau}\}^T_{\tau=1}$ for each task.
The hyper-embedding is fed to a hypernetwork $h(.)$ parameterized by $\theta_{h}$, which generates task-specific
parameters $\Delta \theta = h(\theta_{h}, I_{\tau})$ for other networks.
In the multitask training, the hypernetwork is trained to capture the shared knowledge across tasks.
It enables positive transfer among related domains and tasks, and mitigates the catastrophic forgetting problems to some extent.
In this paper, we use a simple linear projection layer as the hypernetwork that takes hyper-embeddings $I_{\tau}$ as input, and outputs weights for subset networks of PLMs.

\section{Proposed Method}

We aim to integrate a unified hypernetwork-based
parameter-efficient transfer learning method into a multi-task transformer model.
In other word, we insert
the parameters generated by the hypernetworks $\Delta \theta$ into the layer and attention blocks of PLMs.
During training, we only update the hypernetwork parameters $\theta_{h}$ with hyper-embeddings
$\{I_{\tau}\}^T_{\tau=1}$ and parameters in layer normalization,
while the remaining model parameters in $\theta$ are fixed as in the Equation~\ref{equ:loss_2}.
\begin{equation}
\begin{split}
\mathcal{L}_{\text{total}}  & = \sum^T_{\tau=1} \sum_{(x^i_{\tau},y^i_{\tau})\in\mathcal{D}_{\tau}} \mathcal{L}_{\text{task}} (\Delta \theta, \theta, x^i_{\tau}, y^i_{\tau}) \\
                            & = \sum^T_{\tau=1} \sum_{(x^i_{\tau},y^i_{\tau})\in\mathcal{D}_{\tau}} \mathcal{L}_{\text{task}} (I_{\tau}, \theta_h, \theta, x^i_{\tau}, y^i_{\tau})
\end{split}
,
\label{equ:loss_2}
\end{equation}

We next describe the detailed hyper-embedding $I_{\tau}$ and which modules of PLMs to insert the parameters $\Delta \theta$ generated by hypernetworks, to achieve PELT. In our methods, the hyper-embeddings $I_{\tau}$ consists of two: $I^t_{\tau}$ is computed by a task projector network $h^t_I(.)$ for each individual task, and $I^v_{\tau}$ is computed by a visual projector network $h^v_I(.)$ for each image. We will mostly introduce the hyper-embedding $I^t_{\tau}$, and $I^v_{\tau}$ is used in a similar parallel manner.

\subsection{Hyper-Embeddings for PELT}\label{subsec:hyperembeds}

Considering a flexible parameterization of task-specific
parameters for $L$ layers of transformer,
we introduce a set of layer id embeddings $\mathcal{I} = \{l_i\}^L_{i=1}$, and 
block type embeddings $\mathcal{B} = \{ b_j \}^5_{j=1}$,
which specify the position where
the parameters $\Delta \theta$ are inserted to.
The five block positions to insert $\Delta \theta$ include the self-attention, feed-forward block of an encoder, and the self-attention, cross-attention, and feed-forward block in a decoder.
The layer id and block type embeddings together determine the exact targeted transformer block,
and are also used as input to the hypernetwork.

Then, we compute a hyper-embedding $I^t_{\tau} \in \mathbb{R}^{d_I}$ for each individual task via a task projector network $h^t_I(.)$.
It is a multi-layer perceptron consisting of two feed-forward layers and a ReLU non-linearity:
\begin{equation}
\label{equation_hyper_embedding}
I^t_{\tau} = h^t_I (z_{\tau}, l_i, b_j) ,
\end{equation}
where
the task projector network $h^t_I(.)$ learns a suitable compressed hyper-embedding from a concatenation of task embeddings $z_{\tau} \in \mathbb{R}^{d_t}$, layer id embeddings $l_i \in \mathbb{R}^{d_t}$, and block type embeddings $b_j \in \mathbb{R}^{d_t}$.
In this way, this hypernetwork is able to produce distinct weights for tuning each task, and each transformer block at each layer.

\subsection{HyperPrefix: Incorporate with Prefix-tuning}\label{subsec:hyperprefix}

Prefix-tuning~\cite{DBLP:conf/acl/LiL20} prepends a number of task-specific trainable prefix vectors to the parameters of multi-head attention (\textit{i.e.,} keys and values) at each transformer layer.
In the original implementation, the prefix vectors of each attention block are reparameterized by a two-layer feed-forward network:
\begin{equation}
P = W_{\text{up}} \phi ( W_{\text{down}} E ) ,
\end{equation}
where $P \in \mathbb{R}^{d \times N}$, $W_{\text{down}} \in \mathbb{R}^{d_{\text{mid}} \times d}$, $W_{\text{up}} \in \mathbb{R}^{ d \times L \times 2 \times d_{\text{mid}}}$,
$N$ denotes the prefix length, $d$ denotes the dimension size of PLMs and $E \in \mathbb{R}^{d \times N}$ is a randomly initialized embedding matrix. For prefix-tuning, the block type $\mathcal{B}$ has three: the self-attention block of encoder, the self-attention block and cross-attention block of decoder.
Therefore, this original method indeed introduces quite a large number of parameters in the finetuning phase, due to the separate embedding matrix $E$ and feed-forward networks at each block.

In our method, we extend the dimension for different embeddings to match the prefix length $N$, \textit{i.e.,} $z_{\tau} \in \mathbb{R}^{N \times d_t}$, $l_i \in \mathbb{R}^{N \times d_t}$, $b_j \in \mathbb{R}^{N \times d_t}$, and then compute the hyper-embedding $I^t_{\tau} \in \mathbb{R}^{N \times d_I}$. We finally employ a hypernetwork $h^t_P(.)$ with trainable parameters $\theta_{h^t_P}$, to project $I^t_{\tau}$ to prefix vectors $P^t \in \mathbb{R}^{N \times d}$, named \textit{HyperPrefix}:
\begin{equation}
P^t = h^t_P (\theta_{h^t_P}, I^t_{\tau}) .
\end{equation}
In this way, since the hyper-embedding has already perceived information of which block at which layer, the number of hypernetwork parameters, \textit{i.e.,} $\theta_{h^t_P}$, can be largely reduced to $\mathbb{R}^{d_I \times d}$.
We further explain the relationship between prefix-tuning and hypernetwork in the Appendix~\ref{sec:prefix_hyper}.

\subsection{HyperPELT: Incorporate with Adapter}\label{subsec:hyperPELT}

To further capture knowledge across tasks and transfer to others, we follow the adapter-tuning~\cite{DBLP:journals/corr/abs-2110-04366}, and input the hyper-embedding to a hypernetwork for generating the weights in adapters. %
As depicted in Figure~\ref{fig:method}, 
we introduce a hypernetwork-based adapter layer with a trainable scaled parameter $\lambda$, which is inserted parallelly with feed-forward blocks, named HyperPELT.
This task-conditioned adapter layer $A_{\tau}$
consists of a down-projection,
$W^{\tau}_{\text{down}} \in \mathbb{R}^{d_{\text{mid}} \times d}$, 
GeLU non-linearity, and up-projection
$W^{\tau}_{\text{up}} \in \mathbb{R}^{d \times d_{\text{mid}}}$,
where $h$ is the input dimension, $d_{\text{mid}}$ is the bottleneck dimension for the adapter layer, and $x$ is the input hidden state, mathematically defined as:
\begin{equation}
A^t_{\tau}(x) = \lambda \ \text{LN} (W^{\tau}_{\text{up}} \mathrm{GeLU}(W^{\tau}_{\text{down}}x)) + x ,
\label{eq:adapter}
\end{equation}
We generate adapter weights $(W^{\tau}_{\text{up}}, W^{\tau}_{\text{down}})$ through a hypernetwork $h^t_A(.)$:
\begin{equation}
(W^{\tau}_{\text{up}}, W^{\tau}_{\text{down}}) := h^t_A (\theta_{h^t_A}, I^t_{\tau}) .
\end{equation}
Note that in Section \ref{subsec:hyperprefix}, we use the prefix length $N$ as the dimension for hyper-embeddings. We utilize an adaptive pooling operation on hyper-embeddings to adjust the dimension for adapter hypernetwork.
Note that due to we extend the dimension of the components of hyper-embeddings in the last section, we utilize an adaptive pooling operation for hyper-embeddings to adjust the dimension for adapter hypernetwork.

\subsection{VL-HyperPELT: Incorporate with Visual Modality}\label{subsec:VL-hyperPELT}
Transferring pretrained language models to V\&L tasks~\cite{DBLP:conf/icml/ChoLTB21}, with updating the whole parameters is inefficient.
Thus, it motivates us to employ parameter-efficient tuning approaches to extend language models for V\&L tasks.
We follow~\citet{DBLP:conf/icml/ChoLTB21} to unify V\&L tasks as a text generation problem.
As illustrated in Fig.~\ref{fig:method}, we use CLIP~\cite{DBLP:conf/icml/RadfordKHRGASAM21} (parameterized by $\theta^v$) with a trainable visual projection layer (parameterized by $\theta^{v \rightarrow I_v}$), which projects the visual representation to the identical dimension of task embedding, \textit{i.e.,} $z_{\tau} \in \mathbb{R}^{N \times d_t}$. Then we feed this visual representation $v$ to a visual projector network $h^v_I(.)$, whose architecture is same to the mentioned task projector network $h^t_I(.)$.
In this way, we learn the visual hyper-embeddings $I^v_{\tau} \in \mathbb{R}^{d_I}$.
Finally, taking this visual-specific hyper-embeddings as input, we use a visual hypernetwork $h^v(.)$ to generate new visual-specific parameters to different modules in PLMs.

Similar to the Section~\ref{subsec:hyperprefix}~\&~\ref{subsec:hyperPELT}, the incorporation of  visual-specific parameters to PLMs are the same as task-specific ones, \emph{e.g.,} used as prefix vectors via a prefix hypernetwork $h^v_P(.)$ and adapter weights via an adapter hypernetwork $h^v_A(.)$. We name it VL-HyperPELT. Note that $I^v_{\tau}$ is also calculated as the Equation~\ref{equation_hyper_embedding} via replacing task embedding $z_{\tau}$ by visual representation $v$. As illustrated in Fig.~\ref{fig:method}, the hyper-embeddings $I^t$ and $I^v$ separately produce weights for parallel adapters. In addition, their produced prefix vectors are appended together for multi-head attention modules.

\begin{table*}[t]
\centering
 \resizebox{\linewidth}{!}{%
\begin{tabular}{l|cc|cccccccc|c}
\toprule
Model & \tabincell{c}{\#Total\\params} & \tabincell{c}{\#Trained \\ params/task} & CoLA & SST-2 & MRPC & QQP & STS-B & MNLI & QNLI & RTE & Avg \\
\midrule
\midrule
\multicolumn{4}{l}{\textbf{\textit{Single-Task Training}}} \\
\midrule
T5$_{\text{BASE}}$ $\dagger$
    & 8.0$\times$       & 100\%     & 54.85 & 92.19 & \underline{88.18}/\underline{91.61} & \underline{91.46}/\underline{88.61} & \underline{89.55}/\underline{89.41} & \underline{86.49} & 91.60 & 67.39 & 84.67 \\
Adapters $\dagger$
    & 1+8$\times$0.01   & 0.87\%    & \underline{59.49} & \underline{93.46} & \underline{88.18}/91.55 & 90.94/88.01 & 87.44/87.18 & 86.38 & \underline{92.26} & \underline{68.84} & \underline{84.88} \\
\midrule
\midrule
\multicolumn{4}{l}{\textit{\textbf{Multi-Task Training}}} \\
\midrule
T5$_{\text{BASE}}$ $\dagger$
    & 1.0$\times$       & 12.5\%    & 54.88 & 92.54 & \underline{90.15}/\underline{93.01} & \underline{91.13}/\underline{88.07} & 88.84/88.53 & 85.66 & 92.04 & \underline{75.36} & 85.47 \\
Adapters $\dagger$
    & 1.07$\times$      & 0.82\%    & 61.53 & 93.00 & \underline{90.15}/92.91 & 90.47/87.26 & 89.86/89.44 & \underline{86.09} & \underline{93.17} & 70.29 & 85.83 \\
HYPERFORMER++ $\dagger$
    & 1.02$\times$      & 0.29\%    & \underline{63.73} & \underline{94.03} & 89.66/92.63 & 90.28/87.20 & \underline{90.00}/\underline{89.66} & 85.74 & 93.02 & \underline{75.36} & \underline{86.48} \\

\midrule
T5$_{\text{BASE}}$ $\spadesuit$
    & 1.0$\times$       & 12.5\%    & 55.58 & 93.00 & 87.50/90.97 & 88.96/85.54 & 88.96/88.40 & 83.62 & 91.47 & 81.25 & 85.02 \\
Prefix-tuning $\clubsuit$
    & 1.14$\times$      & 1.72\%    & 56.67 & \underline{93.92} & 89.42/92.57 & \underline{90.59}/87.37 & \underline{89.49}/\underline{89.34} & 85.23 & \underline{93.17} & 79.17 & 86.09 \\
MAMAdapters $\clubsuit$
    & 1.15$\times$      & 2.96\%    & 56.53 & 93.58 & \underline{91.35}/\underline{93.96} & 90.58/\underline{87.53} & 88.89/88.76 & \underline{85.98} & 92.77 & 81.94 & 86.53 \\
HYPERFORMER++ $\spadesuit$
    & 1.02$\times$      & 0.23\%    & \underline{58.02} & 93.69 & 91.34/93.84 & 90.42/87.25 & 88.94/88.68 & 85.41 & 92.51 & \underline{83.33} & \underline{86.68} \\
\midrule
HyperPrefix
    & 1.01$\times$      & 0.15\%    & 63.01 & \underline{93.46} & \underline{90.38}/\underline{93.10} & \underline{90.49}/87.27 & \underline{89.88}/\underline{89.71} & 85.21 & \underline{92.88} & 77.78 & 86.65 \\
HyperPELT
    & 1.02$\times$      & 0.24\%    & \underline{65.96} & 93.23 & 89.42/92.31 & 90.48/\underline{87.54} & 89.15/89.07 & \underline{85.35} & 92.79 & \underline{82.64} & \underline{\textbf{87.09}} \\
\bottomrule
\end{tabular}
}
\caption{
Performance of all models on the GLUE tasks.
For each method, we report the total number of parameters across all tasks and the number of parameters that are trained for each task as a multiple and proportion respectively of the baseline single-task \textit{T5} model.
For MNLI, we report accuracy on the matched validation set.
For MRPC and QQP, we report accuracy and F1.
For STS-B, we report Pearson and Spearman correlation coefficients.
For CoLA, we report Matthews correlation.
For all other tasks, we report accuracy.
$\dagger$: Results from the implementation of \citet{DBLP:conf/acl/MahabadiR0H20},
$\spadesuit$: Our re-implementation of \cite{DBLP:conf/acl/MahabadiR0H20},
$\clubsuit$: We implement the methods of \citet{DBLP:conf/acl/LiL20} and \citet{DBLP:journals/corr/abs-2110-04366} on top of \textit{T5}.
}
\label{tab:glue}
\end{table*}

\begin{table}[t]
\centering
 \resizebox{1.0\linewidth}{!}{%
\begin{tabular}{c | cccc | cc}
\toprule
Dataset & \rotatebox{60}{\# Samples} & \rotatebox{60}{T5$_{\text{BASE}}\dagger$} & \rotatebox{60}{HYPERFORMER++$\dagger$} & \rotatebox{60}{HyperPELT} & \rotatebox{60}{Prompt-tuning} & \rotatebox{60}{HyperPELT TaskEmbed}  \\

\midrule
\multicolumn{7}{c}{Natural Language Inference} \\
\midrule

\multirowcell{5}[0pt][l]{CB}
& 4     & 57.78$_{\pm\text{10.9}}$  & 60.74$_{\pm\text{16.66}}$ & \textbf{77.86$_{\pm \text{6.9}}$}  & 81.43$_{\pm \text{5.2}}$  & \textbf{85.71$_{\pm \text{3.2}}$} \\
& 16    & 77.04$_{\pm\text{7.2}}$   & 76.29$_{\pm\text{4.45}}$  & \textbf{82.14$_{\pm \text{3.9}}$}  & 84.29$_{\pm \text{4.3}}$  & \textbf{86.43$_{\pm \text{1.4}}$} \\
& 32    & 80.00$_{\pm\text{7.6}}$   & 81.48$_{\pm\text{6.2}}$   & \textbf{83.57$_{\pm \text{3.6}}$}  & 85.71$_{\pm \text{3.2}}$  & \textbf{87.14$_{\pm \text{1.7}}$} \\
& 100   & 85.93$_{\pm\text{5.4}}$   & 87.41$_{\pm\text{2.96}}$  & \textbf{90.71$_{\pm \text{2.9}}$}  & 87.14$_{\pm \text{1.7}}$  & \textbf{87.86$_{\pm \text{1.7}}$} \\
& 250   & 85.19$_{\pm\text{4.7}}$   & 89.63$_{\pm\text{4.32}}$  & \textbf{91.43$_{\pm \text{2.8}}$}  & 88.57$_{\pm \text{3.5}}$  & \textbf{89.57$_{\pm \text{1.4}}$} \\

\midrule
\multicolumn{7}{c}{Question Classification} \\
\midrule

\multirowcell{7}[0pt][l]{TREC}
& 4     & 28.11$_{\pm\text{5.9}}$   & \textbf{28.85$_{\pm\text{6.9}}$}   & 24.08$_{\pm \text{1.7}}$  & \textbf{21.76$_{\pm \text{1.7}}$}  & 18.88$_{\pm \text{0.6}}$ \\
& 16    & 40.08$_{\pm\text{12.6}}$  & \textbf{49.40$_{\pm\text{9.5}}$}   & 48.96$_{\pm \text{5.0}}$  & \textbf{34.00$_{\pm \text{13.8}}$} & 19.20$_{\pm \text{0.7}}$ \\
& 32    & 62.49$_{\pm\text{6.2}}$   & 68.94$_{\pm\text{7.5}}$   & \textbf{69.28$_{\pm \text{3.8}}$}  & \textbf{58.24$_{\pm \text{18.1}}$} & 20.88$_{\pm \text{1.8}}$ \\
& 100   & 87.79$_{\pm\text{0.7}}$   & 88.42$_{\pm\text{1.7}}$   & \textbf{88.96$_{\pm \text{1.3}}$}  & \textbf{84.96$_{\pm \text{3.8}}$}  & 22.64$_{\pm \text{2.8}}$ \\
& 500   & 93.57$_{\pm\text{1.3}}$   & 94.78$_{\pm\text{1.4}}$   & \textbf{96.16$_{\pm \text{0.6}}$}  & \textbf{87.52$_{\pm \text{3.1}}$}  & 23.76$_{\pm \text{1.6}}$ \\
& 1000  & 95.5$_{\pm \text{0.9}}$   & 96.72$_{\pm \text{1.3}}$  & \textbf{97.04$_{\pm \text{0.6}}$}  & \textbf{88.32$_{\pm \text{3.0}}$}  & 24.40$_{\pm \text{0.8}}$ \\
& 2000  & 96.87$_{\pm \text{1.3}}$  & 96.92$_{\pm \text{0.9}}$  & \textbf{97.20$_{\pm \text{0.4}}$}  & \textbf{91.68$_{\pm \text{3.1}}$}  & 24.64$_{\pm \text{0.8}}$ \\

\midrule
\multicolumn{7}{c}{Question Answering} \\
\midrule

\multirowcell{7}[0pt][l]{BoolQ}
& 4     & 50.49$_{\pm\text{11.1}}$  & 48.03$_{\pm\text{4.8}}$   & \textbf{71.52$_{\pm \text{0.1}}$}  & 64.69$_{\pm \text{8.8}}$  & \textbf{69.28$_{\pm \text{4.1}}$} \\
& 16    & 56.50$_{\pm\text{7.1}}$   & 50.21$_{\pm\text{7.9}}$   & \textbf{73.66$_{\pm \text{1.8}}$}  & \textbf{73.44$_{\pm \text{1.8}}$}  & 71.52$_{\pm \text{0.2}}$ \\
& 32    & 58.43$_{\pm\text{4.9}}$   & 58.37$_{\pm\text{3.7}}$   & \textbf{73.80$_{\pm \text{0.6}}$}  & \textbf{75.16$_{\pm \text{0.8}}$}  & 72.64$_{\pm \text{0.4}}$ \\
& 100   & 60.10$_{\pm\text{2.4}}$   & 62.03$_{\pm\text{2.0}}$   & \textbf{75.70$_{\pm \text{0.8}}$}  & \textbf{75.57$_{\pm \text{1.0}}$}  & 73.59$_{\pm \text{0.4}}$ \\
& 500   & 66.49$_{\pm\text{1.2}}$   & 70.04$_{\pm\text{1.4}}$   & \textbf{76.37$_{\pm \text{0.5}}$}  & \textbf{75.82$_{\pm \text{1.1}}$}  & 73.73$_{\pm \text{0.7}}$ \\
& 1000  & 69.01$_{\pm \text{1.1}}$  & 72.35$_{\pm \text{1.7}}$  & \textbf{76.71$_{\pm \text{1.6}}$}  & \textbf{76.91$_{\pm \text{0.3}}$}  & 74.33$_{\pm \text{0.6}}$ \\
& 2000  & 71.58$_{\pm \text{0.8}}$  & 74.94$_{\pm \text{0.6}}$  & \textbf{77.81$_{\pm \text{0.8}}$}  &\textbf{ 77.11$_{\pm \text{0.6}}$}  & 74.53$_{\pm \text{0.1}}$ \\

\midrule
\multicolumn{7}{c}{Sentiment Analysis} \\
\midrule

\multirowcell{7}[0pt][l]{IMDB}
& 4     & 77.23$_{\pm\text{3.0}}$   & \textbf{81.77$_{\pm\text{1.8}}$}   & 80.68$_{\pm \text{0.4}}$  & 50.68$_{\pm \text{0.9}}$  & \textbf{56.41$_{\pm \text{0.3}}$} \\
& 16    & 82.74$_{\pm\text{1.7}}$   & 84.06$_{\pm\text{0.7}}$   & \textbf{84.86$_{\pm \text{3.8}}$}  & 54.83$_{\pm \text{1.8}}$  & \textbf{59.57$_{\pm \text{1.3}}$} \\
& 32    & 83.42$_{\pm\text{1.0}}$   & 84.64$_{\pm\text{0.4}}$   & \textbf{85.40$_{\pm \text{0.4}}$}  & 58.70$_{\pm \text{7.9}}$  & \textbf{62.53$_{\pm \text{1.5}}$} \\
& 100   & 84.58$_{\pm\text{0.6}}$   & 84.74$_{\pm\text{0.4}}$   & \textbf{90.50$_{\pm \text{0.4}}$}  & 76.31$_{\pm \text{11.1}}$ & \textbf{86.68$_{\pm \text{0.2}}$} \\
& 500   & 84.99$_{\pm\text{0.3}}$   & 86.00$_{\pm\text{0.2}}$   & \textbf{91.78$_{\pm \text{0.3}}$}  & 89.36$_{\pm \text{0.4}}$  & \textbf{89.41$_{\pm \text{0.3}}$} \\
& 1000  & 85.50$_{\pm \text{0.1}}$  & 86.37$_{\pm \text{0.4}}$  & \textbf{92.19$_{\pm \text{0.3}}$}  & 89.68$_{\pm \text{0.7}}$  & \textbf{90.15$_{\pm \text{0.2}}$} \\
& 2000  & 86.01$_{\pm \text{0.2}}$  & 86.60$_{\pm \text{0.1}}$  & \textbf{92.54$_{\pm \text{0.1}}$}  & 90.08$_{\pm \text{0.7}}$  & \textbf{91.78$_{\pm \text{0.1}}$} \\

\midrule
\multicolumn{7}{c}{Paraphrase Detection} \\
\midrule

\multirowcell{7}[0pt][l]{PAWS}
& 4     & 53.89$_{\pm\text{3.6}}$   & 55.58$_{\pm\text{7.5}}$   &\textbf{ 59.58$_{\pm \text{0.4}}$}  & 54.54$_{\pm \text{4.1}}$  & \textbf{56.47$_{\pm \text{0.2}}$} \\
& 16    & 54.18$_{\pm\text{1.0}}$   & 72.71$_{\pm\text{1.1}}$   & \textbf{73.66$_{\pm \text{4.4}}$}  & 55.49$_{\pm \text{0.6}}$  & \textbf{56.92$_{\pm \text{0.4}}$} \\
& 32    & 55.23$_{\pm\text{3.2}}$   & 73.39$_{\pm\text{2.1}}$   & \textbf{74.41$_{\pm \text{1.2}}$}  & 55.93$_{\pm \text{0.8}}$  & \textbf{58.99$_{\pm \text{0.3}}$} \\
& 100   & 71.51$_{\pm\text{2.4}}$   & 78.24$_{\pm\text{2.1}}$   & \textbf{79.84$_{\pm \text{1.3}}$}  & 57.56$_{\pm \text{0.9}}$  & \textbf{59.35$_{\pm \text{1.4}}$} \\
& 500   & 82.81$_{\pm\text{1.0}}$   & 86.30$_{\pm\text{1.1}}$   & \textbf{87.31$_{\pm \text{1.1}}$}  & 63.94$_{\pm \text{0.9}}$  & \textbf{64.71$_{\pm \text{0.3}}$} \\
& 1000  & 85.67$_{\pm\text{0.7}}$   & 89.12$_{\pm\text{0.5}}$   & \textbf{90.42$_{\pm \text{0.3}}$}  & 65.55$_{\pm \text{7.6}}$  & \textbf{66.80$_{\pm \text{0.4}}$} \\
& 2000  & 88.33$_{\pm\text{0.6}}$   & 90.87$_{\pm\text{0.3}}$   & \textbf{91.79$_{\pm \text{0.6}}$}  & 67.17$_{\pm \text{1.4}}$  & \textbf{68.67$_{\pm \text{0.6}}$} \\

\bottomrule
\end{tabular}
}
\caption{
Few-shot domain transfer results of five different tasks
averaged across 5 seeds.
We compute accuracy for all tasks and datasets.
$\dagger$: Results from the paper of \citet{DBLP:conf/acl/MahabadiR0H20}.
HyperPELT and HyperPELT TaskEmbed are respectively fine-tuning hypernetworks with all hyper-embeddings and only task embedding in the few-shot learning.
}
\label{tab:fewshot}
\end{table}

\begin{table*}
\centering
\resizebox{.9\linewidth}{!}{
\begin{tabular}{l|c|c|ccc|c|cccc}
\toprule
& \multirowcell{2}[0pt][c]{Trained \\ Params (\%)} & VQAv2 & \multicolumn{3}{c|}{VQA Karpathy test} & GQA & \multicolumn{4}{c}{CoCo Caption} \\
& & test-std & in-domain & out-domain & overall & test-dev & B@4 & M & C & S \\
\midrule
\midrule
\multicolumn{4}{l}{\textbf{\textit{Single-Task Training}}} \\
\midrule
VL-T5 $\dagger$
    & 100\%     & 70.3 & 71.4 & 13.1 & 67.9 & 60.0 & 34.6 & 28.8 & 116.1 & 21.9 \\
\midrule
\midrule
\multicolumn{4}{l}{\textbf{\textit{Multi-Task Training}}} \\
\midrule
VL-T5 $\dagger$
    & 100\%     & - & - & - & 67.2 & \underline{58.9} & - & - & 110.8 & - \\
CLIP-T5 $\dagger$
    & 100\%     & - & - & - & 67.3 & 56.5 & - & - & \underline{113.1} & - \\
VL-Adapter $\dagger$
    & 7.98\%    & - & - & - & \underline{67.6} & 56.2 & - & - & 111.8 & - \\
\midrule
CLIP-T5 $\spadesuit$
    & 100\%     & \underline{69.8} & \underline{70.8} & \underline{17.4} & \underline{66.8} & \underline{59.6} & \underline{32.4} & 27.1 & \underline{108.5} & \underline{20.4} \\
VL-Adapter $\spadesuit$
    & 7.16\%    & 69.4 & 70.0 & 16.4 & 65.9 & 57.6 & 31.4 & \underline{27.2} & 105.6 & 20.1 \\
VL-HyperPELT
    & 6.62\%    & 69.6 & 70.3 & 16.8 & 66.3 & 57.9 & 32.1 & 27.0 & 108.2 & 20.1 \\
\bottomrule
\end{tabular}
}
\caption{
Experimental results on popular Vision-Language banchmarks.
We report vqa-score for VQA, gqa-score for GQA
and various metrics for image captioning (B@4: BLEU@4, M: METEOR, C: CIDEr, S: SPICE).
$\dagger$: Results from the paper of \citet{DBLP:conf/icml/ChoLTB21,sung2021vladapter},
$\spadesuit$: Our re-implementation of \citet{sung2021vladapter}.
Following~\citet{DBLP:conf/icml/ChoLTB21}, we use VQA Karpathy split, which splits the VQA dataset into 605,102 / 26,729 / 26,280 image and question pairs separately as the train/validation/test set to evaluate VQA tasks in a generative manner.
}
\label{tab:vl}
\end{table*}

\begin{table}[t]
\centering

 \resizebox{.9\linewidth}{!}{
\begin{tabular}{l|ccccc}
\toprule
Dataset & \rotatebox{44}{\# Samples}
& \rotatebox{44}{CLIP-T5} & \rotatebox{44}{VL-Adapter} & \rotatebox{44}{VL-HyperPELT} \\ %

\midrule
\multicolumn{5}{c}{Knowledge-based VQA} \\
\midrule
\multirowcell{7}[0pt][l]{OKVQA}
& 4     & 32.65$_{\pm\text{0.4}}$   & 31.83$_{\pm\text{1.1}}$   & \textbf{33.25$_{\pm\text{0.5}}$} \\ %
& 16    & 33.68$_{\pm\text{1.0}}$   & 31.86$_{\pm\text{0.6}}$   & \textbf{34.72$_{\pm\text{0.6}}$} \\ %
& 32    & 33.87$_{\pm\text{1.1}}$   & 32.07$_{\pm\text{0.9}}$   & \textbf{34.86$_{\pm\text{0.3}}$} \\ %
& 100   & 34.27$_{\pm\text{0.1}}$   & 33.03$_{\pm\text{1.6}}$   & \textbf{34.99$_{\pm\text{0.9}}$} \\ %
& 500   & 34.43$_{\pm\text{2.1}}$   & 33.35$_{\pm\text{0.8}}$   & \textbf{35.56$_{\pm\text{1.1}}$} \\ %
& 1000  & 34.59$_{\pm\text{1.1}}$   & 34.57$_{\pm\text{0.2}}$   & \textbf{35.72$_{\pm\text{0.4}}$} \\ %
& 2000  & 34.62$_{\pm\text{0.9}}$   & 34.87$_{\pm\text{0.6}}$   & \textbf{35.86$_{\pm\text{0.6}}$} \\ %

\midrule
\multicolumn{5}{c}{Visual Entailment} \\
\midrule

\multirowcell{7}[0pt][l]{SNLI-VE}
& 4     & 34.77$_{\pm\text{2.2}}$   & 38.75$_{\pm\text{4.3}}$   & \textbf{39.94$_{\pm\text{0.9}}$} \\ %
& 16    & 38.55$_{\pm\text{2.6}}$   & 46.67$_{\pm\text{1.7}}$   & \textbf{47.86$_{\pm\text{1.2}}$} \\ %
& 32    & 39.89$_{\pm\text{3.7}}$   & 51.69$_{\pm\text{2.4}}$   & \textbf{53.40$_{\pm\text{0.9}}$} \\ %
& 100   & 49.97$_{\pm\text{1.8}}$   & 55.05$_{\pm\text{1.9}}$   & \textbf{57.69$_{\pm\text{0.8}}$} \\ %
& 500   & 59.01$_{\pm\text{1.6}}$   & 60.58$_{\pm\text{1.1}}$   & \textbf{61.97$_{\pm\text{1.1}}$} \\ %
& 1000  & 60.49$_{\pm\text{0.6}}$   & 62.83$_{\pm\text{0.4}}$   & \textbf{63.01$_{\pm\text{0.6}}$} \\ %
& 2000  & 62.29$_{\pm\text{0.9}}$   & 64.22$_{\pm\text{1.1}}$   & \textbf{65.67$_{\pm\text{0.7}}$} \\ %

\bottomrule
\end{tabular}
}
\caption{
Few-shot domain transfer results of two different V\&L tasks
averaged across 5 seeds.
We report the vqa-score on OKVQA validation split, and the accuracy on SNLI-VE test-P split.
}
\label{tab:vl_fewshot}
\end{table}

\section{Experimental Setup}

\subsection{Datasets}
Our framework is evaluated
on the GLUE benchmark~\cite{DBLP:conf/iclr/WangSMHLB19} in terms of natural language understanding.
This benchmark covers multiple tasks of paraphrase detection (MRPC, QQP), sentiment classification (SST-2), natural language inference (MNLI, RTE, QNLI), and linguistic acceptability (CoLA).
The original test sets are not publicly available, and following~\citet{DBLP:conf/iclr/0007WKWA21}, for datasets fewer than 10K samples (RTE, MRPC, STS-B, CoLA), we split the original validation set into two halves, one for validation and the other for testing.
For other larger datasets, we randomly split 1K samples from the training set as our validation data and test on the original validation set.

In addition, we evaluate the few-shot domain transfer performance on four tasks and datasets:
1) the natural language inference (NLI) datasets
CB and 2) the question answering (QA) dataset BoolQ from SuperGLUE~\cite{DBLP:conf/nips/WangPNSMHLB19};
3) the sentiment analysis datasets IMDB~\cite{DBLP:conf/acl/MaasDPHNP11};
and
4) the paraphrase detection dataset PAWS~\cite{DBLP:conf/naacl/ZhangBH19}.
For CB and BoolQ, since the test set is not available, we split the validation set into two halves, one for validation and the other for testing.
For
IMDB, since the validation set is not available, we similarly split the test set to form validation. For PAWS, we report on the original test set.

To evaluate our framework on V\&L tasks, we experiment on four datasets
COCO~\citep{DBLP:conf/eccv/LinMBHPRDZ14},
VQA~\citep{DBLP:conf/cvpr/GoyalKSBP17},
VG-QA~\citep{DBLP:journals/ijcv/KrishnaZGJHKCKL17} and
GQA~\citep{DBLP:conf/cvpr/HudsonM19}.
We further evaluate our framework on
three datasets for multi-modal few-shot transfer learning:
OKVQA~\citep{DBLP:conf/cvpr/MarinoRFM19};
SNLI-VE~\citep{DBLP:journals/corr/abs-1811-10582}.

\subsection{Implementation Details}
Our models are built on \emph{T5$_{\text{BASE}}$}~\citep{DBLP:journals/jmlr/RaffelSRLNMZLL20}~\footnote{https://huggingface.co/t5-base} and use the same tokenizer of \emph{T5} to tokenize text inputs.
We set $N=49$, $d=d_t=768$, $d^{\text{mid}}_I=128$, $d_I=64$ for all the experiments.
Following the training strategies from~\citet{DBLP:journals/jmlr/RaffelSRLNMZLL20}, we fine-tune all models with a constant learning rate of 0.001, use $2^{18}=262144$
steps in all experiments with batch size of 128 and sample tasks via the conventional temperature-based sampler, with temperature $T=2$, i.e.,
sample corresponding task proportional to $p^{1/T}_{\tau}$, where $p_{\tau} = \frac{N_{\tau}}{\sum^{T}_{i=1} N_{\tau}}$ and $N_{\tau}$ is the number of training samples for the $\tau$-th task.
We did not experiment with other complex sampling strategies or tuning of $T$.
For the experiments under multi-task training settings, we save a checkpoint every 1000 steps and report results on a single checkpoint with the highest average validation performance across all tasks.

For evaluating our framework on vision-language scenarios,
we follow~\citet{DBLP:conf/icml/ChoLTB21}
to convert V\&L tasks to a text generation format.
We use \textit{ResNet101} as our vision encoder, and initialize it with CLIP~\cite{DBLP:conf/icml/RadfordKHRGASAM21}~\footnote{https://github.com/openai/CLIP} pretrained weights.
Input images are resized to 224 $\times$ 224 for the memory efficiency.
We extract the 7 $\times$ 7 grid features produced by the last convolutional layer.
The percentage of updated parameters is also reported as one metric for approach efficiency, and we do not take visual encoder into computation since it is frozen in our experiments.
We count the number of tunable parameters and list the input-output formats of each task in the Appendix~\ref{sec:parameters} and~\ref{sec:format}.

\section{Results and Analysis}

We design a series of experiments for pure language and V\&L tasks on both multi-tasking and few-shot scenarios to verify the effectiveness of our proposed framework compared to existing methods. 

\subsection{Results on the GLUE Benchmark}

We conduct experiments on GLUE for both single- and multi-task settings. 
As shown in Table~\ref{tab:glue}, we also report the total number of parameters and trainable parameters for all models.
Our methods are built on \textit{T5$_{\text{BASE}}$} which contains 12 layers and 222M parameters. 
For adapter-related methods, we experiment with reduction factors of $r = 32$ \cite{DBLP:conf/acl/MahabadiR0H20},
where $r = \frac{d}{d_{\text{mid}}}$.
We follow the original \textit{T5} implementation \cite{DBLP:journals/jmlr/RaffelSRLNMZLL20}, which is slightly different from \textit{HYPERFORMER++}~\cite{DBLP:conf/acl/MahabadiR0H20}.
In our re-implementation of \textit{T5$_{\text{BASE}}$} and \textit{HYPERFORMER++}, we prepend task-prefix tokens to the textual input sequence as the original \textit{T5} does.
We also experiment with generating weights for layer normalization via a hypernetwork.
However, we find that it makes no difference on performances.

For \textit{Prefix-tuning}~\cite{DBLP:conf/acl/LiL20} and \textit{MAMAdapter}~\cite{DBLP:journals/corr/abs-2110-04366}, their original implementation is single-task training on \textit{BART}~\cite{DBLP:conf/acl/LewisLGGMLSZ20}.
To make a fair comparison to other baselines, we apply their methods to \textit{T5} in a multi-task training setting.~\footnote{For adapting Prefix-tuning from BART to T5, a noteworthy point is that since they use different position embedding, i.e., absolute position embedding for BART and relative position embedding for T5, it is necessary to manually concatenate all-zero vectors on the relative position bias of each layer in T5.}
For each model, we share the parameters of both prefix vectors and adapter weights across multitasks.

Overall, our \textit{HyperPELT} method obtains the best performance with less trainable parameters.
Compared to the single-task \emph{Adapters} that finetunes all the introduced parameters in adapters, our method yields a significant improvement by 2.21\% with much fewer trainable parameters, which illustrates the effectiveness of our proposed multi-task training framework.

In multi-task training, the proposed hypernetwork-based prefix-tuning strategy, \textit{e.g., HyperPrefix}, decreases the number of trainable parameters (\textit{e.g.}, 1.01$\times$ of \textit{HyperPrefix} vs. 1.14$\times$ of \textit{Prefix-tuning}), while achieves a better performance at the same time (\textit{e.g.}, 86.65\% of \textit{HyperPrefix} vs. 86.09\% of \textit{Prefix-tuning}). 
It is noticeable that the number of trainable parameters per task is 11$\times$ fewer than \textit{Prefix-tuning}. 

\textit{HyperPELT} obtains a superior performance over \textit{HyperPrefix}, and the main reason lies in that we further combine the hypernetwork-based adapters and add them to the feedforward layers in a parallel manner.
In this way, the average performance is further enhanced (+0.44\%) by involving a small number of parameters (0.09\% parameters per task).
The comparison to \textit{MAMAdapter} shows that using hypernetwork to tune each transformer block and learn the shared knowledge across multitasks leads to an improvement.

\subsection{Few-shot Domain Transfer} \label{subsec:fewshot}

We use the above models trained on GLUE as reported in Table~\ref{tab:glue}, and evaluate them on the test set of five different tasks after being few-shot finetuned on each target training data.
Following~\citet{DBLP:conf/acl/MahabadiR0H20}, we use
the task embedding respectively trained on the most similar GLUE task for initialization, i.e., MNLI for CB,
QNLI for QA, SST-2 for sentiment analysis, and QQP for paraphrase detection.
As suggested by~\citet{DBLP:journals/corr/abs-2105-11447} and \citet{DBLP:conf/emnlp/ZhaoS21}, we randomly select the same number of samples from training and validation set, making it a reasonable few-shot scenario.
Checkpoints are selected via early stopping on the selected validation set, and the stopping metric is the default metric for each task.

In the first three columns of Table~\ref{tab:fewshot}, we show the results of full fine-tuning of \textit{T5$_{\text{BASE}}$}, 
\textit{HYPERFORMER++} (finetuning both hypernetworks and task embeddings) and our proposed \textit{HyperPELT}. Overall, our method achieves the best performance in the few-shot tasks.

For the tasks of CB and BoolQ from SuperGLUE, even though the backbone \textit{T5} was previously trained on the train sets of these two, the performance of all methods differs a lot.
The two baselines still do not work with very few samples, like 4 and 16 samples, while our method is significantly better than them.
Therefore, we assume that the two baselines suffer from catastrophic forgetting problems to some degree during multi-task training.
In contrast, our proposed \textit{HyperPELT} works effectively on these two tasks.
We speculate that the reason might be the use of hypernetworks on both prefix-tuning and adapter-tuning modules of transformer.
We leave this exploration to our future work.

Besides, in the last two columns of Table~\ref{tab:fewshot}, we show the results of \textit{Prompt-tuning}~\cite{DBLP:journals/corr/abs-2104-08691} and fine-tuning only the task embedding in our \textit{HyperPELT}.
Note that in this comparison, we keep the same trainable parameters between these two methods, \textit{i.e.,} $\mathbb{R}^{N \times d_t}$, where $N$ denotes the prompt length in \textit{Prompt-tuning} method.
Our \textit{HyperPELT TaskEmbed} mostly achieves a comparable or even better performance than \textit{Prompt-tuning} except TREC task, where the task format and verbalizer are quite different from previous tasks. In this case, only tuning the very limit parameters of our model with very few samples is unable to quickly transfer.  In comparison, \textit{HYPERFORMER++}~\cite{DBLP:conf/acl/MahabadiR0H20} claims only fine-tuning the task embedding in their model achieves low performance in the few-shot learning and they have not reported any result. Our framework with only fine-tuning the task embedding, though not perfect on all tasks, may still present a new promising parameter-efficient way towards few-shot learning with an extreme low number of tunable parameters \cite{DBLP:journals/corr/abs-2110-08207}.

\subsection{Results on the Vision-Language Benchmarks}
Next, we move to the experiments of applying the proposed hypernetwork-based parameter-efficient training framework to V\&L tasks.
We compare to the pre-trained and full fine-tuning \textit{VL-T5}~\cite{DBLP:conf/icml/ChoLTB21}, and other adapter-based methods built on top of \textit{T5}, \textit{i.e.,} \textit{CLIP-T5} and \textit{VL-Adapter}~\cite{sung2021vladapter} in the multi-task training setting.

The results and the number of trainable parameters are reported in Table~\ref{tab:vl}.
To match the same amount of trainable parameters as \textit{VL-Adapter},
we experiment with the reduction factors $r = 16$,
where $r = \frac{d}{d_{mid}}$.
Since 
the test dataset is slightly different from~\citet{sung2021vladapter} and their checkpoint is not avaliable at this time, we re-implement \textit{CLIP-T5} and \textit{VL-Adpater}.
Compared to which, our method achieves a comparable performance with fewer number of trainable parameters (\textit{e.g.}, 7.16\% of \textit{VL-Adapter} vs. 6.62\% of \textit{VL-HyperPELT}).

To our best knowledge, we are the first to employ the visual modality to tune the very few parameters of different transformer blocks, instead of normally inserting image patch tokens to the input sequence.
Experimental results evidence the effectiveness of our novel approach, thus providing a new perspective on how to extend the multi-modality capability on top of PLMs. It is to use the features from different modalities as the input of hypernetwork to generate parameters for modules in PLMs, instead of as a part of input sequence to accomplish the multimodal tasks. One advantage in our approach is still keeping the original maximum text input length, since no other modalities such as visual and audio features occupy it. It is promisingly useful in document-level and text-heavy tasks such as multimodal summarization~\cite{zhang2021unims}.

We believe the resulting performance might be even better with a more complex design combination of methods across tuning task-specific and visual-specific parameters in PLMs,
but we leave this exploration in future work.

\subsection{Multimodal Few-shot Learning}

We further use the models trained on V\&L tasks as reported in Table~\ref{tab:vl_fewshot} and evaluate them on the test set after few-shot fine-tuning on OKVQA~\citep{DBLP:conf/cvpr/MarinoRFM19} and SNLI-VE~\citep{DBLP:journals/corr/abs-1811-10582}.
For OKVQA, since there is no test set, we split its original validation set into two halves, one for validating and the other for testing. For SNLI-VE, we use its validation set for validating, and test-P set for testing and reporting results. We follow the methods in Section~\ref{subsec:fewshot} to select samples, and report results in Table~\ref{tab:vl_fewshot}.

Compared with the full parameter fine-tuning, \textit{i.e.,} \textit{CLIP-T5}, and the other baseline \textit{VL-Adapter}, our method achieves the best performance with smaller variances in this multimodal few-shot learning setting.
We find that \textit{VL-Adapter} is inferior to \textit{CLIP-T5} when with fewer samples (\textit{e.g.}, fewer than 500) on the OKVQA dataset. 
The reason may be that there exists a lot of out-domain knowledge and complex image content in OKVQA, which makes it more challenging for parameter-efficient \textit{VL-Adapter} to achieve accurate prediction. 
In other words, the small number of samples are not enough to train the introduced randomly initialized parameters in \textit{VL-Adapter}.

However, our approach can still tackle with fewer samples. We use the hypernetwork to generate trainable parameters in adapters and multi-head attention, as well as directly integrating image features into attention modules in the form of prefix tuning vectors. We believe such method, though training less parameters, can still capture knowledge across tasks and transfer them in a few-shot setting.
It is also worth noting that for the used five random seeds, the variance of our method is generally smaller than \textit{VL-Adapter}, which indicates that our method is more robust in this few-shot learning scenario.

\section{Discussion and Conclusion}

In this paper, we propose a unified parameter-efficient tuning framework for multitasks, particularly on both pure language and vision-and-language (\textit{i.e.,} V\&L) tasks. On the one hand, we use a hypernetwork to reduce the scale of trainable parameters of existing adapter-tuning and prefix-tuning modules. 
On the other hand, for the V\&L tasks, we directly integrate the image features into the multi-head attention in the form of prefix vectors, which further reduces the number of trainable parameters for processing visual input. 
Extensive experiments on pure language and V\&L tasks demonstrate the superiority of our proposed framework in both multi-tasking and few-shot settings. In the future, we plan to explore more combination of methods across tuning task-specific and visual-specific parameters for different modules in a pretrained language model.

%\bibliography{example_paper,mybib}
\bibliography{mybib}
\bibliographystyle{icml2022}

\appendix
\onecolumn

\section{The Connection Between Prefix-tuning and Hypernetwork}\label{sec:prefix_hyper}

Prefix tuning prepends $N$ tunable prefix vectors to the keys and values of the multi-head attention at every layer.
Specifically, two sets of prefix vectors $P_k, P_v \in \mathbb{R}^{N \times d}$ are concatenated with the original key $K$ and value $V$.
Then multi-head attention is performed on the new prefixed keys and values.
The computation of head$_i$:
\begin{equation}
\mathrm{head}_i = \mathrm{Attn}(x W^{l_i}_q, \mathrm{concat}(P^{l_i}_k, CW^{l_i}_k), \mathrm{concat}(P^{l_i}_v, CW^{l_i})) .
\label{eq:head}
\end{equation}

\cite{DBLP:journals/corr/abs-2110-04366} derives an equivalent form of Eq.~\ref{eq:head} and provide an alternative view of prefix tuning:
\begin{equation}
\begin{split}
\mathrm{head} & = \mathrm{Attn}(x W_q, \mathrm{concat}(P_k, CW_k), \mathrm{concat}(P_v;CW_v)) \\
& = \mathrm{softmax} (x W_q \mathrm{concat}(P_k, CW_k)^{\top}
[
\begin{array}{cc}
     & P_v \\
     & CW_v
\end{array}
] \\
& = (1 - \lambda(x)) \mathrm{softmax}(x W_q W_k^{\top} C^{\top}) CW_v + \lambda(x) \mathrm{softmax}(x W_q P_k) P_v \\
& = \underbrace{(1 - \lambda(x)) \mathrm{Attn}(x W_q, CW_k, CW_v)}_{\text{standard attention}} + \underbrace{\lambda(x) \mathrm{Attn}(x W_q, P_k, P_v)}_{\text{indendent of C}} ,
\end{split}
\label{eq:view}
\end{equation}
where $\lambda(x)$ is a scalar that represents the sum of normalized attention weights on the prefixes:
\begin{equation}
\lambda(x) = \frac{ \sum_i \mathrm{exp}(x W_q P_k^{\top})_i }{ \sum_i \mathrm{exp}(x W_q P_k^{\top})_i + \sum_j \mathrm{exp} (x W_q W_k^{\top} C^{\top})_j }
\end{equation}
Eq.~\ref{eq:view} gives an alternative view of prefix tuning that essentially applies a position-wise modification to the original head attention output $h$ through linear interpolation.
Eq.~\ref{eq:view} can be rewrited by define $W_1 = W_q P_k^{\top}$, $W_2 = P_v$, $f = \mathrm{softmax}$:
\begin{equation}
h \leftarrow  (1 - \lambda(x))h + \lambda(x) f (x W_1) W_2,
\end{equation}
which reaches a very similar form to the adapter function in Eq.~\ref{eq:adapter}, except that prefix tuning is performing weighted addition while the adapter one is unweighted.
This view of prefix tuning allows for abstraction of prefix tuning as a plug-in module like adapters.
Therefore, we can approximately regard the prefix vectors as the weights of the model, which can be generated when combined with hypernetworks.

\begin{table*}[b]
\centering
\begin{tabular}{lc}
\toprule
Method          & Number of Tunable Parameters \\
\midrule
Prompt Tuning   & $N \times d$ \\
Prefix Tuning   & $N \times d + (1 + 2 \times L) \times d_{\text{mid}} \times d \times B_{\text{attn}}$ \\
Adapter         & $2 \times d_{\text{mid}} \times d \times (B_{\text{attn}} + B_{\text{ffn}}) \times L$ \\
MAM Adapter     & $N \times d + (1 + 2 \times L) \times d_{\text{mid}} \times d \times B_{\text{attn}} + 2 \times d_{\text{mid}} \times d \times B_{\text{ffn}} \times L$ \\
\midrule
HYPERFORMER++   & $(N + B_{\text{attn}} + B_{\text{ffn}} + L) \times d_t + d_t \times d^{\text{mid}}_I + d^{\text{mid}}_I \times d_I + 2 \times d_I \times (d_{\text{mid}} \times d)$ \\
HyperPrefix     & $(N + B_{\text{attn}} + L) \times d_t + d_t \times d^{\text{mid}}_I + d^{\text{mid}}_I \times d_I \times d_I \times d$ \\
HyperPELT       & $(N + B_{\text{attn}} + B_{\text{ffn}} + L) \times d_t + d_t \times d^{\text{mid}}_I + d^{\text{mid}}_I \times d_I + 2 \times d_I \times d + 2 \times d_I \times (d_{\text{mid}} \times d)$ \\
\bottomrule
\end{tabular}
\caption{
Number of tunable parameters of various parameter-efficient tuning methods with T5 models.
}
\label{tab:params}
\end{table*}

\section{Number of Tunable Parameters}\label{sec:parameters}
Following~\citet{DBLP:journals/corr/abs-2110-04366}, to simplify the computation of tunable parameters, we compute the sum of parameter used in one encoder layer and one decoder layer as the parameter overhead of one single layer of the pre-trained encoder-decoder model.
T5 has an encoder-decoder structure that has $L$ layers.
Each layer has $B_{\text{attn}}$ blocks and $B_{\text{ffn}}$ blocks.
For the encoder-decoder models like T5, $B_{\text{attn}} = 3$: the encoder self-attention block, the decoder self-attention block and the decoder cross-attention block and $B_{\text{ffn}} = 2$: encoder feed-forward block and decoder feed-forward block.
For modifications applied at the attention blocks, the number of tunable parameters is computed by $\theta_{\text{attn}} = \theta^{\text{attn}}_{\text{W}} \times B_{\text{attn}} \times L$, where $\theta^{\text{attn}}_{\text{W}}$ denotes the number of parameters
used for one attention sub-layer.
Similarly, the number of tunable parameters for the FFN sub-layers is computed by $\theta_{\text{ffn}} = \theta^{\text{ffn}}_{\text{W}} \times B_{\text{ffn}} \times L$.
Finally, the total number of tunable parameters for prefix tuning and adapter variants
is $\theta = \theta_{\text{attn}} + \theta_{\text{ffn}}$ as applicable.
Using
T5 as an example, we present the number of parameters used by several representative methods throughout our paper in Tab.~\ref{tab:params}.

\section{Input-output formats}\label{sec:format}

\begin{table*}[t]
\centering
\begin{tabular}{l|l|l}
\toprule
Task    & Input Text & Target Text \\
\midrule
\multicolumn{3}{l}{\textbf{\textit{GLUE Tasks}}} \\
\midrule
CoLA    & cola sentence: [sentence]                             & acceptable/unacceptable \\
SST-2   & sst2 sentence: [sentence]                             & positive/negative \\
MRPC    & mrpc sentence1: [sentence1] sentence2: [sentence2]    & equivalent/not\_equivalent \\
QQP     & qqp question1: [question1] question2: [question2]     & duplicate/not\_duplicate \\
STS-B   & stsb sentence1: [sentence1] sentence2: [sentence2]    & 0.0 - 5.0\\
MNLI    & mnli hypothesis: [hypothesis] premise: [premise]      & entailment/neutral/contradiction \\
QNLI    & qnli question: [question] sentence: [sentence]        & entailment/not\_entailment \\
RTE     & rte sentence1: [sentence1] sentence2: [sentence2]     & entailment/not\_entailment \\
\midrule
\multicolumn{3}{l}{\textbf{\textit{Few-shot Tasks}}} \\
\midrule
CB      & cb hypothesis: [hypothesis] premise: [premise]        & entailment/neutral/contradiction \\
TREC    & trec question: [question]                             & DESC/ENTY/ABBR/HUM/NUM/LOC \\
BoolQ   & boolq question: [question] context: [context]         & True/False \\
IMDB    & imdb sentence: [sentence]                             & positive/negative \\
PAWS    & paws sentence1: [sentence1] sentence2: [sentence2]    & equivalent/not\_equivalent \\
\midrule
\multicolumn{3}{l}{\textbf{\textit{Vision-Language Tasks}}} \\
\midrule
COCO    & caption:                  & [caption] \\
VQA     & vqa question: [question]  & [answer] \\
GQA     & gqa question: [question]  & [answer] \\
\midrule
\multicolumn{3}{l}{\textbf{\textit{Vision-Language Few-shot Tasks}}} \\
\midrule
OKVQA   & okvqa question: [question]    & [answer] \\
SNLI-VE & snli-ve premise: [premise]    & entailment/neutral/contradiction \\
\bottomrule
\end{tabular}
\caption{
Input-output formats for NLU and Vision-Language tasks.
Following~\citet{DBLP:journals/jmlr/RaffelSRLNMZLL20,DBLP:conf/icml/ChoLTB21}, we use different prefixes (such as “cola sentence:”, “vqa question:”) for questions from different datasets.
}
\label{tab:format}
\end{table*}

As shown in Tab.~\ref{tab:format}, we formulate the input text and labels from each task to the corresponding text, and we learn these different tasks by predicting target text with the language modeling objective in Eq.~\ref{equ:loss_2}.

\end{document}